# Transfer Learning for Material Classification using Convolutional Networks

Patrick Wieschollek     Hendrik P.A. Lensch
Tübingen University

## Abstract

*Material classification in natural settings is a challenge due to complex interplay of geometry, reflectance properties, and illumination. Previous work on material classification relies strongly on hand-engineered features of visual samples. In this work we use a Convolutional Neural Network (convnet) that learns descriptive features for the specific task of material recognition. Specifically, transfer learning from the task of object recognition is exploited to more effectively train good features for material classification. The approach of transfer learning using convnets yields significantly higher recognition rates when compared to previous state-of-the-art approaches. We then analyze the relative contribution of reflectance and shading information by a decomposition of the image into its intrinsic components. The use of convnets for material classification was hindered by the strong demand for sufficient and diverse training data, even with transfer learning approaches. Therefore, we present a new data set containing approximately 10k images divided into 10 material categories. Images in this data set are captured in the wild, i.e., they differ in illumination and view (close-shot and object-level view).*

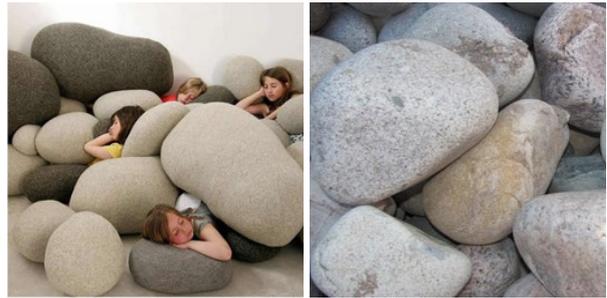

Figure 1. While the main object in both scenes can be considered as "stone" it is obvious that their material is different.

## 1. Introduction

In order to interact with the world around we constantly need to reason about the material of objects and the material properties. Humans have a clear intuitive concept about how materials will look like in different environments and have a strong association of the connection between visual reflection and other physical properties. Take Figure 1 as an example. Humans can immediately distinguish the deformable cloth from the solid rock, based on visual appearance alone. One can easily infer properties like softness, toxicity, stickiness from visual samples, real objects or images, even without exactly knowing the material category [6]. The underlying inference principles are yet largely unknown. Learn to predict the material category or material properties of given visual samples would allow automatically controlled machines to interact with objects in an appropriate way, e.g., dosing the spent amount of power when grabbing an object. As such it is a fundamental problem of computer vision.

The appearance of objects and materials varies strongly with the surrounding environment and the resulting illumination. It is possible that humans use visual traits like properties of shading and specular highlights to judge a material sample. Changes of shading and reflectance are systematically bound to the laws of optical physics, i.e., they cannot change in an arbitrary way. This additional information can help a machine learning approach to discover dependencies between shading, reflectance, and the properties of a visual sample. More precisely, limiting the amount of available information during the learning process by providing either shading or reflectance information, it is possible to identify in detail which traits are useful to decide which material category is depicted.

Instead of only judging local visual traits like texture patches or parts of geometric shapes, the additional understanding of a scene comprises context information, which can help to conclude material categories (compare Figure 1).

Convolutional Neural Networks (convnets) have demonstrated state-of-the-art performance on challenging visual classification tasks such as object recognition [12] and hand-written digit classification [18]. Reasons for their excellent performance are mainly their high-capacity nature, the use of parallel computing enviroments, and the availability of large training sets[1]. A convnet applies non-linear filter stages to images to map the pixel information into a feature space. A filter stage is a combination of an image convolution with a filter, with a subsequent non-linearity like $\tanh(\cdot)$, down-sampling the result by max-pooling/average-pooling, and normalization of the output. During the training phase a convnet optimizes its parameters by each given input. The learned features can be fed into a classifier such as a Multi-Layer-Perceptron (MLP) or a Support-Vector-Machine (SVM) to categorize the material within an image.

---

[1]The ILSVRC 2014 challenge provides 456567 high-resolution images

Despite the common assumption that training and test data have to be in the same feature space and underlie the same distribution, this is not true in many real-world problems. In addition, the human visual system handles many visual impressions for diverging tasks. Hence, it is more likely that the human visual system uses the same method to solve different visual recognition tasks. This transfer of knowledge is known as *transfer learning* [16]. We apply this concept to transfer the structure and weights of a convnet trained for object classification and augment it to boost the performance of material classification.

The main of this paper contributions are

- A material classification framework based on a Convolutional Neural Network with significantly improved performance compared to previous state-of-the-art approaches.
- The application of transfer learning from object classification to material classification in order to reduce the required training data set.
- An analysis of the contribution effects of shading and reflectance on material classification.
- A new large data set for for visual material samples, containing approximately 10k images for 10 material categories.

## 2. Related Work

Learning a high level representation of material categories is considered to be mostly independent from object or texture classification. It is assumed that information about the geometry of the object or the texture can be misleading [14, 1], since the same object can be made of different materials while at the same time different materials can share the same texture.

Therefore in order to build a discriminative model for material classification, previous work used hand-engineered conceptualized features for material recognition in a combined approach consisting of *bag-of-words* and *augmented LDA* [14] or *SVM* [1] as a classifier. The authors of [14, 1] extracted features like color patches, SIFT features, curvature of edges and raw rgb patch strides parallel and orthogonal to image edges. To consider the micro-structure of materials, a subset of these features were extracted from the residual image, i.e., the difference between the plain image and its result after bilateral filtering. All these features can be combined into a discriminative model for material recognition. Performing model selection by evaluating different feature combinations on the test set yields the former state-of-the-art result, i.e., an accuracy of $57.1\%$ on the Flickr Material Database (FMD) [20]. This benchmark set contains 100 images in close-shot view and object-level view for 10 high level material categories (fabric, foliage, glass, leather, metal, paper, plastic, stone, water and wood). Another approach [19] makes exhaustive use of additional accurate pixel-wise labels. Instead of learning the material category directly, they learn attributes of $32 \times 32$ pixel patches, e.g., metallic, soft, smooth, liquid, rough. However this labeling is not practical for larger data sets and only achieves an accuracy of $49.2\%$ on the FMD. While an advantage of this approach is the possibility to partition an image concerning the material category, its does not use the context between local regions and the complete image. Both approaches utilize binary masks describing whether a pixel in the image belongs to the material that should be classified or not. In fact, the abstract problem can be formulated as mapping a four channel image (rgb and mask) to a label. The authors of [15] employed methods from texture classification for material classification by performing *random projections* of features from a high dimensional space into a low dimensional space for classification. This achieves an accuracy of $48.2\%$ on the FMD. Interestingly, this agrees with the reported performance of random filters in convnets for feature extraction in the limit of very small data sets [9]. These approaches used standard SVMs to classify a bunch of features and share the drawback that it is difficult to compute posterior probabilities [7].

Previous work has demonstrated that a convnet, once having learned good low- and mid-level features in some problem domains of image classification, can also be used for such classification task, too. Examples are learning the image style [11] or classifying Chinese characters [3]. The authors of [17] showed that a convnet model trained on the *ILSVRC13* challenge archives very good performances on several classification tasks and was able to obtain many state-of-the-art or competitive results in *bird subclassification*, *flowers recognition*, *human attribute detection*, *Oxford buildings retrieval* or *object instance retrieval* among others. To apply transfer learning by convnets, one can fix initial layers in the net and just finetune a classifier from the extracted features. Consequently, this method can improve the performance of convnets even on relatively small training sets like the standard FMD for material classification.

Intrinsic image decomposition is an ill-posed inverse problem with many applications in computer graphics and vision [24]. The objective is to explain a change of color

---
for training.

between pixels concerning the change of shading $s_{ij} \in \mathbb{R}$ and the change of albedo (reflectance) $\mathcal{R}_{ij} \in \mathbb{R}^3$ by assuming the factorization of an image by $\mathcal{I}_{ij} = s_{ij}\mathcal{R}_{ij}$ for all pixels identified by their position $(i,j)$. It can be used for stylization and recolorization of a given image/video or to change visible material properties like specular highlights [24]. Although the basic idea is to create a probabilistic model specifying a conditional probability over the components of reflectance and shading in an image

$$p(s, \mathcal{R}) \propto \exp(-E(s, \mathcal{R}|\mathcal{I})),$$

various formulations of the energy function $E(\cdot)$ have been proposed [8, 24, 2].

## 3. Novel Material Data Set

The Flickr Material Database can be considered the current benchmark data set for existing material recognition algorithms. Its shortcomings are the small number of labeled examples (100 images per category) as well as the presence of multiple materials in an image, for which only a single material label is available. Recently, [2] introduced a new data set that contains 105076 images with annotated masks for some material categories. However, these images are mostly indoor scenes and only small parts of the images contain the annotated material. A BTF Material Database [23] uses independently acquired material appearances and illuminations to *synthesize* photo-realistic training examples. We create a new data set containing approximately 10k real-world images with minimum size $256 \times 256$ pixels from arbitrary places in an uncontrolled environment. While the ImageNet data set would offer more annotated images per category, these images are categorized in objects categories rather than in material categories (compare Figure 2) and usually contain multiple objects and materials in one image.

Collecting images from the internet was done by creating a list of words with 174 queries in multiple languages that mostly belong to a specific material category and using the Google-Image-Search to query for images tagged with words from this list that match the following criteria (filtered by Google-Image-Search):

1. The size of the image is at least $400 \times 300$ pixel, such that we are able to crop the region that contains the material sample.
2. The image has to be a color image.
3. The type of the image has to be a photo without obvious clip-arts.

From each query result all entries were downloaded in high resolution for further processing. From these $124,308$ downloaded images we removed exact duplicates and blank images. We also manually sorted out images that do not contain large enough regions containing a single material, images showing multiple overlapping materials, manipulated photos, and photo compositions. The remaining images were trimmed to remove single color strides at each side. Images with a size less than $400 \times 300$ pixels were rejected. From the remaining images we manually extracted regions that contain the material that corresponds to the related tags. If the image did not contain such a region, we removed it from our data set. Finally, all larger images were resized such that the smallest dimension is at least 384 pixels.

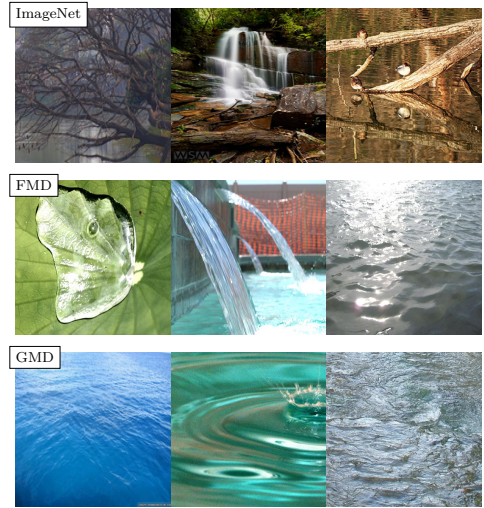

Figure 2. Examples labeled with "water" from different data sets. Former data sets like the Flickr Material Database (FMD) or the ImageNet data set contain images consisting of multiple different material. Using these examples to learn to predict a material category can be misleading. We introduce a novel data set consisting of samples containing a single material.

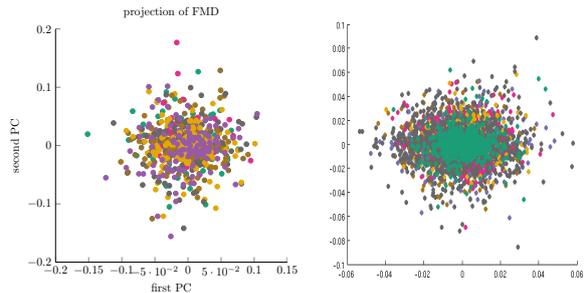

Figure 3. The projection on the first two principal components (PCs) of the Leung-Malik filter bank are shown for all images of the 10 categories of the FMD (left) and our data set (right).

To verify our data set does not have examples which are classifiable in an easy way, we projected random patches of size $60 \times 60$ pixels of our data set and the FMD onto the first and second principal component of the *Leung-Malik* filter bank results (see Figure 3) similar to [14]. In addition, the accuracy (compare Figure 4) of our classifier trained on the novel material data set and tested on both test sets (complete FMD and our new data set) reveals that our data set is comparable to the FMD within the meaning of severity. The differences between both data sets in the category *water* are caused by the appearance of multiple materials in images of the FMD (compare Figure 2).

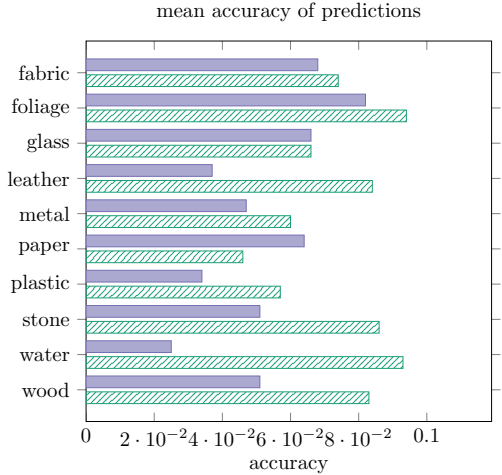

Figure 4. To compare the complexity of both data sets FMD (■) and GMD (▨) we run our best deep convnet and plot the per-category accuracy.

For future benchmarks we randomly split our data set "Google Material Database" (GMD) into 6742 images for training, 2000 images for validation (200 images of each category) and 1000 images for testing (100 images of each category). The data set will be made available for future research.

## 4. Material Classification

The training process of convnets includes carefully choosing many hyper-parameters, like the number of layers and the size of the receptive field of a single neurons by specifying the size of convolution filters. While a simple and shallow convnet model avoids overfitting of data due to the reduced number of trainable parameters, a deep convnet containing many parameters can lead to highly specialized parameters depending on few training examples. As a simple baseline approach in applying convnets in material recognition, a modified *part* of the well studied *LeNet*-layout [13] in object recognition is used.

A common technique to enable training of deep networks on small data sets is to inject knowledge from other computer vision tasks by deploying weights and parameters from a pretrained network to the new one. To pursue the idea of knowledge-transfer, we used a publicly available model [10] trained on the task of recognize objects in the *ImageNet ILSVRC 2012 challenge*.

### 4.1. Vanilla Convnet Approach

Learn to recognize material categories just from a small amount of training data requires a shallow convnet. This simple approach applies a simple convnet with one filter to find the most informative feature in classifying visual examples. Therefore this simple convnet consists of a modified version of the first filter stage in the fashion of the *LeNet*-layout. Following the notation of [9] we write $xF_{\text{CSG}}^{n\times m}$ for $x$ convolution filters of size $n \times m$, $R_{\text{relu}}$ for the rectification layer, $P_M^{n\times m}$ for the max-pooling layer of size $n \times m$, $N$ for the local contrast normalization layer, $xI$ for a fully connected layer of size $x$, dropout$_x$ for a dropout layer [22] with dropout ratio $x$ as a regularizer and $xS$ for a softmax-layer with $x$ outputs.

The layout of the vanilla convnet can be written as

$$96F_{\text{CSG}}^{11\times 11} - R_{\text{relu}} - P_M^{6\times 6} - \text{dropout}_{0.5}$$
$$-4096I - R_{\text{relu}} - 4096I - 10S.$$

Hereby the convnet learns features picked from patches of size $11 \times 11$ pixels and down-samples these information before using a MLP as a classifier. We did not apply any preprocessing steps like data normalization, e.g., subtract the mean image of training set or mask the image. Note, that some images may be misleading due to the occurrence of multiple materials in a single image having only one label. This approach already outperforms all known results from single features [1, 14] classified by SVM or LDA (see Table 1).

### 4.2. Application of Transfer Learning

The concept of the *LeNet*-layout has been proven to perform well in several visual recognitions tasks like object recognition [12] and hand-written digit classification [18]. Considering the size of the former benchmark data set (FMD) the only practicable way to apply a deep convnet is to reduce the number of free parameters by just finetuning the MLP part of the convnet meanwhile all other parameters are copied by transfer learning and are kept fixed. More precisely, we use the reference model of Caffe [10] trained on the task of object recognition in the *ImageNet ILSVRC 2012 challenge* and fixed all filter stages as illustrated in Figure 5. If more labeled samples exists then the data set can be split into training samples and a separate validation set to vary and tune the number of fixed layers. We apply the latter approach on the GMD. In detail the layout of the used convnet can be described as five filter stages:

- $96F_{\text{CSG}}^{11\times 11} - R_{\text{relu}} - P_M^{3\times 3} - N$
- $256F_{\text{CSG}}^{5\times 5} - R_{\text{relu}} - P_M^{3\times 3} - N$
- $384F_{\text{CSG}}^{3\times 3} - R_{\text{relu}}$
- $384F_{\text{CSG}}^{3\times 3} - R_{\text{relu}}$

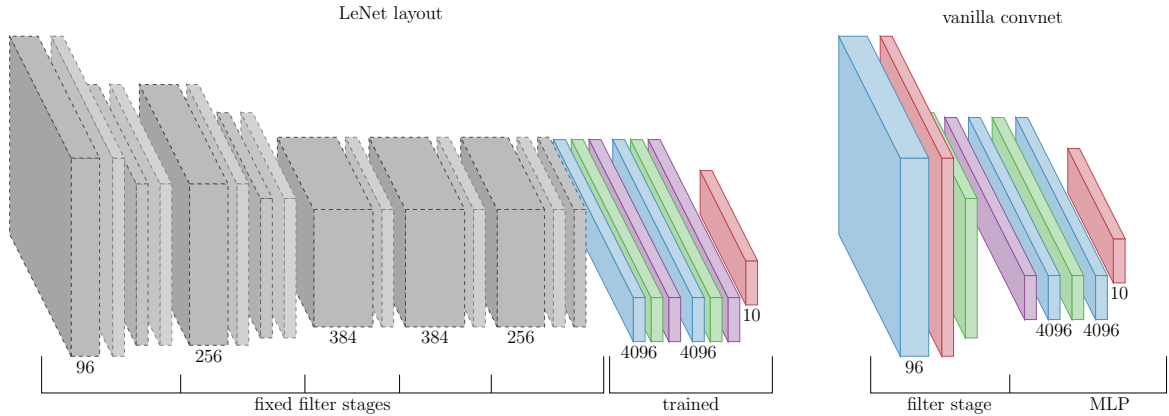

Figure 5. Because of the small amount of obtainable training examples it is difficult to train all parameters of the complete deep convnet. Left: To reduce the number of changeable parameters, the first 5 filter stages (gray dashed) can be trained in advance on different tasks of object recognition and held fixed during training on material recognition. Right: The layout of the shallow convnet that outperforms all single hand-engineered features.

- $384F_{\text{CSG}}^{3\times3} - R_{\text{relu}} - P_M^{3\times3}$

and a MLP with dropout and softmax loss:

- $4096I - R_{\text{relu}} - \text{dropout}_{0.5}$
- $4096I - R_{\text{relu}} - \text{dropout}_{0.5} - 10S$

The layout of a convnet combines features from any location in the image in the higher level layers. It therefore includes features of multiple stages for its inference, which is different to local feature representations such as bag-of-words, where the global reasoning is present only through the final classifier. To enable post-processing the convnet is taught to learn a probability distribution over the material categories by using a softmax-loss layer. The number training samples were increased by randomly crop patches of size $227 \times 277$ pixel from each image as an input for the convnet. The additional information of binary masks were not used.

## 5. Experiments

The evaluation of the deep convnet models were done on normalized rgb images, i.e., the mean image of the training set was subtracted from each sample. The experiments on FMD and GMD will be discussed separately.

### 5.1. Results on the Flickr Material Database

For our convnet approach the FMD data set was split into a test set comprising 200 images (20 randomly chosen images from each category) for testing and into a training set with 800 images. Although usually a data set is split into a training, test and validation set for methods like early-stopping and model selection or tuning hyper-parameters, these experiments do not use a validation set to be comparable to the results reported from the experiments of [14, 1, 21, 4]. The maximum number of iterations was set to $4.5 \cdot 10^5$ for which we found good convergence. The learning rate was initialized to $\eta = 10^{-4}$ and decreased by a factor $0.1$ after every $10^3$ iterations. As an update rule we used *adaptive gradient* [5] with batch size 1. All results of our convnet experiments on the *FMD* are reported in Table 1 and based on the same split of training and test data.

| approach | accuracy |
| --- | --- |
| B-O-W and LDA (cf. [14]) | 44.6% ($\star$) |
| B-O-W and SVM (cf. [1]) | 53.1% ($\star$) |
| kernel descriptors and SVM (cf. [4]) | 54% |
| B-O-W and SVM (cf. [21]) | 55.6% |
| B-O-W and SVM (cf. [21]) | 57.1% ($\star$) |
| random projections (cf. [15]) | 48.2% ($\star$) |
| sparse auto-encoders (cf. [19]) | 49.2% ($\star$) |
| vanilla convnet | 35.7% |
| transfer-learn. convnet (rgb) | **64.0 %** |
| transfer-learn. convnet (shading) | 51.0% |
| transfer-learn. convnet (reflectance) | 54.0% |
| transfer-learn. convnet (reflectance & shading) | 44.0% |

Table 1. Results of approaches trained on the Flickr Material Database only. Accuracies noted by ($\star$) use the information of binary masks which annotate if a pixel belongs to the material that should be classified. A convnet trained on rgb images using transfer learning outperforms all previous reported results significantly. "&" means jointly learned.

### 5.2. Results on the new Material Data Set

One can exploit the size of the new material data set and the existence of the validation set to perform model selec-

| approach | accuracy |
|---|---|
| transfer-lern. (shading, tested on GMD) | 53% |
| transfer-lern. (reflectance, tested on GMD) | 62% |
| transfer-lern. (rgb, tested on GMD) | **74%** |
| transfer-lern. (rgb, tested on FMD) | 52% |

Table 2. Trained on the Google Material Database, the best convnet version via cross-validation has only *one fixed* filter stage and is tested in both data sets. However, shading or reflection information only are not sufficient to achieve good performance. The appearance of many different materials in a single image can misguide the convnet model trained on the GMD which does not contain misleading examples.

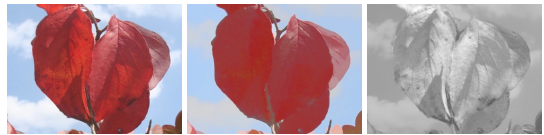

Figure 7. The convnet is trained separately on the original image $\mathcal{I}$ (left), the reflectance information $r(\mathcal{I})$ (middle) and the shading information $s(\mathcal{I})$ (right).

formation, like presenting shading information $s(\mathcal{I})$ or reflectance information $r(\mathcal{I})$ of the image only (see Figure 7). In summary the given information are

- rgb image $\mathcal{I}$ consisting of 3 channels,
- reflectance image $r(\mathcal{I})$ consisting of 3 channels or
- shading image $s(\mathcal{I})$ consisting of 1 channel.

Each image from both data sets was decomposed into an intrinsic image using the publicly available Code of Bell et al. [2]. Figure 8 illustrates the confusion matrices of the same convnet layout trained on different information. These results indicate that predictions relying solely on shading information are likely to misinterpret visual samples that belong to the class of *fabric* as *leather*, *paper*, or *stone*. Other examples where the convnet makes mistakes are: predicting *wood* as *stone*, and *plastic* as *metal*. This is not surprising because these pairs share similar reflection properties, e.g., specular highlights look similar between plastic and metal. Likewise, *wood* and *stone* have a similar diffuse shading. In contrast, using the reflectance information only, the convnet can easily distinguish *foliage*, *stone*, *water*, and *wood* from other categories. In fact these materials can be often classified by just judging the typical color. When providing the rgb representation, the convnet mostly predicts *glass* as *water*. But this representation helps in resolving ambiguities between plastic and other categories.

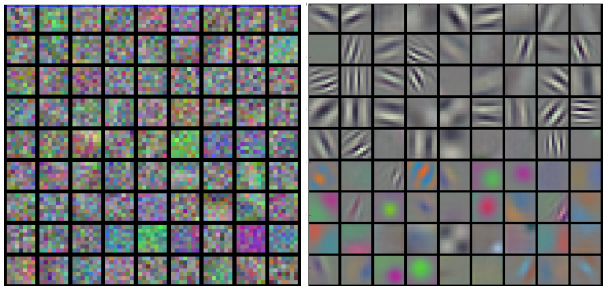

Figure 6. While the first 81 filters from the first layer of the vanilla convnet (left) acts like a color detector containing colored noise, the first 81 filters from the first layer of the convnet model trained on ImageNet (right) combines edge and color detection.

tion via comparing the performance of different models on the same evaluation set. Therefore we tested various versions of the described convnet layout on the GMD by fixing different combination of filter stages. Again the parameters of the layers were initialized by the pretrained model from the ImageNet ILSVRC 2012 challenge. The mutable layers were adapted by finetuning during training on the FMD. Table 2 reports the results from the best performing model on the validation set, i.e., the model were only the first filter stage is kept fixed.

## 6. Discussion

Convolution layers, acting as local filters, have a strong connection to the visual cortex. Figure 6 shows the low level filters from our experiments. The learned filters in the first layer from our vanilla convnet when trained on the FMD contains colored noise, indicating that the convnet mostly acts like a color detector in contrast to the learned filters from the convnet that has been trained on the ImageNet data set only. Note, that this suggests most information about material membership is encoded in the color instead of other sophisticated features.

To further analyze which other traits are useful to successfully classify a visual sample into high level material classes, a convnet can be trained on artificially limited in-

One may ask whether it is sufficient to use shading and reflection information instead of rgb images. For this reason a branched version of the deep convnet in the sense of two instances of the same convnet with a shared softmax layer and fixed filter stages were jointly trained on reflectance information, respectively shading information, at the same time. It is not possible to combine the different modalities successfully as shown in Table 1. In general, it seems that there is a necessity to use more intrinsic information rather than shading and reflection only to support the classification of material. Another possible explanation of the weak results when using shading and reflectance information jointly instead of the rgb representation can be the high depency to one specific intrinsic decomposition. Note that the used state-of-the-art algorithm [2] for intrinsic images relies on several chosen hyper-parameters.

The output of the convnet were trained as the conditional probability distribution over the material categories given the input image. Therefore, in addition to a predicted label the output of the convnet can loosley be interpreted as the confidence of the given prediction. Figure 9 illustrates the confidence of each prediction broken down by the material category and available information. Predictions that rely on the reflectance information only have higher confidence in the categories *water*, *wood* and *stone* than predictions that use the shading information only. Hence, it is likely that the color of theses materials represents an important feature when building a discriminative model. The average confidence when correctly predicting *plastic* is mostly equal between the use of shading or reflectance information. This may be attributed to the highly diverse appearance of plastic in color and shading.

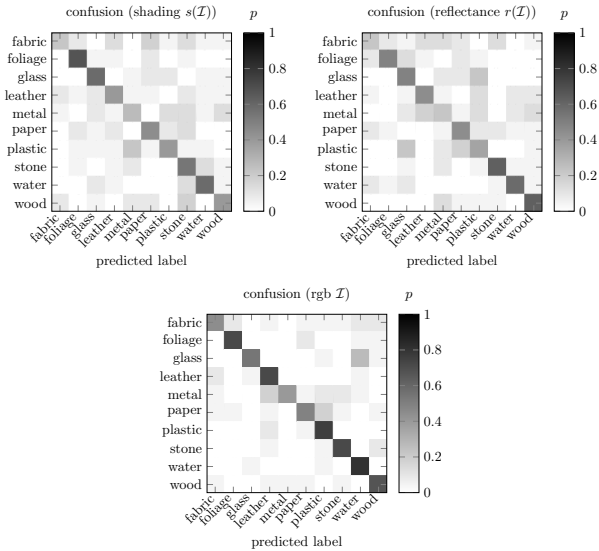

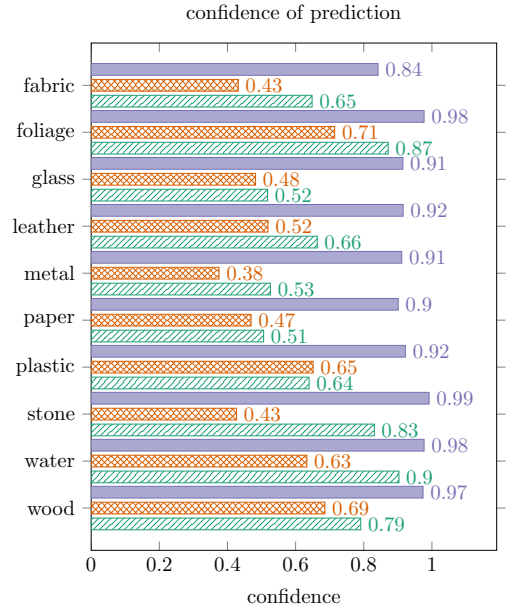

Figure 8. Confusion matrix of the deep discriminative model using transfer learning. Row $n$ is the probability distribution of category $n$ being classified to each category. By changing the representation of the image from rgb information $\mathcal{I}$ to shading $s(\mathcal{I})$ and reflectance information $r(\mathcal{I})$ it is possible to analyze the contribution of each information per category. The category of *foliage* and *water* can be easily recognized in absence of color information. But in contrast, the availability of information about color is crucial to resolve ambiguities between *plastic* and *metal*.

Figure 9. The average confidence of correct predictions is different regarding the available information: reflectance (▨), shading (▨) or rgb (▬). The information about shading and reflectance create equal confidence when correctly recognizing glass.

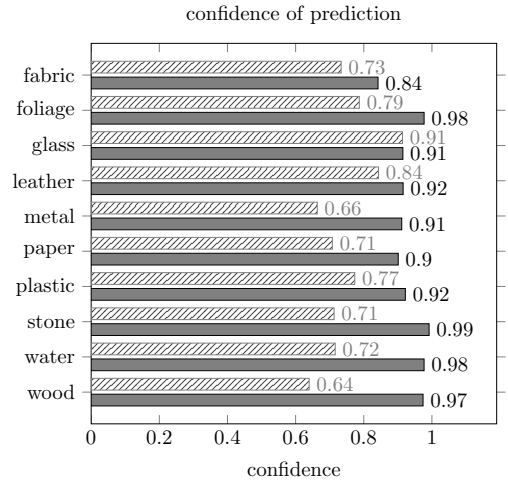

Figure 10. The convnet has less confidence when giving wrong prediction (▨) than in the case of giving correct predictions (▬). The average confidence of correct predictions of the material category *wood* is 0.97 while in rare cases where the convnet is misclassifying other images as wood, the confidence is just 0.64.

It is natural to ask whether the average confidence between correct prediction and wrong prediction differs. Comparing the confidence between correct predictions and wrong predictions (Figure 10) shows that for the category *glass*, the difference in *confidence* between correct and wrong predictions given rgb information is significant. Given an image of glass the convnet has high confidence about its prediction even if the prediction is incorrect. A possible reason may be that the existence of glass can be only indirectly perceived by realizing distorted geometry and shading of the object in the background.

Figure 11 shows wrong predictions with the highest confidence for some categories broken down by the available information.

We did some experiments where we used the information of the available binary mask by replacing pixels by the mean color of the image or simply by setting the corresponding pixels white that do not belong to the target ma-terial. This did not yield into any improvements compared to our results. Probably, this modification removes necessary context between the material sample and the scene it is embedded in.

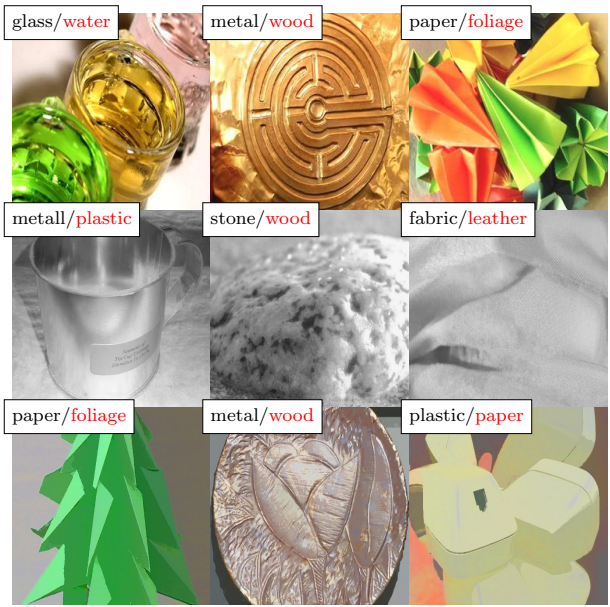

Figure 11. Wrong predictions with highest confidence broken down by the available information for the convnet: rgb information (top row), shading information (middle row) and reflectance information (bottom row). The label glass/water means that the image which belongs to the category glass was predicted as water.

### 6.1. Convnets trained on the GMD

We applied the same experiments from our convnet approach at the FMD on the Google Material Database. The results from Figure 12 confirm that the reflectance information is a good feature for classifying the category *foliage*, *water* and *stone*, while it hard is to distinguish between *plastic* and other categories.

In contrast to the FMD the larger amount of training examples of *leather* in the GMD allows the convnet to isolate this category from the class of *fabric*, although they share many properties. Distinguish *paper* from *foliage* remains a challenging problem even under presence of more training examples.

## 7. Conclusion

The paper applies convolutional networks to the task for material recognition. Even a very simple convnet trained on the Flickr Material Dataset already outperforms all published hand-engineered features in material recognition. As it is not possible to train a deep learning convnet on the small FMD data set we apply transfer learning, making use of parameters trained on the task of object recognition and then retrain fewer parameters on a novel 10 times larger data set we assembled from Google images. The data set will be made available for future research. The resulting deep convent significantly increases the performance on material classification in natural images.

In order to identify the relevance of different visual features, we further train the net on intrinsic images, decomposed in a reflectance and a shading layer. It turns out that different visual features are important for distinguishing different classes of materials as the appearance in each class shows distinct variations of colors or specular highlights.

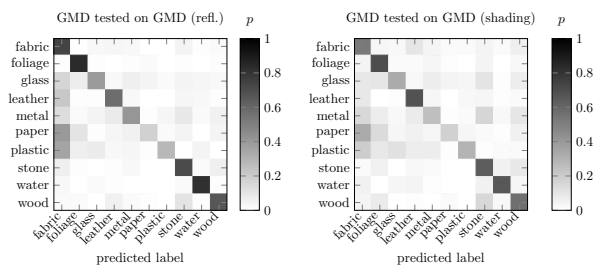

Figure 12. Accuracy of our best performing convnet evaluated at the separate test set.

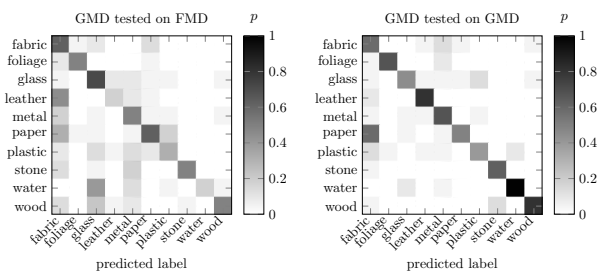

Figure 13. The confusion matrices of our convnets when trained on the GMD (rgb information) an tested on GMD and FMD. We selected the best model evaluated on the separate validation set.

# References

[1] I. Badami. Material recognition: Bayesian interference or svms? In *CESCG*, 2012.

[2] S. Bell, K. Bala, and N. Snavely. Intrinsic images in the wild. *ACM Trans. on Graphics (SIGGRAPH)*, 33(4), 2014.

[3] D. C. Ciresan, U. Meier, and J. Schmidhuber. Transfer learning for latin and chinese characters with deep neural networks. In *IJCNN'12*, pages 1–6, 2012.

[4] L. B. Diane Hu and X. Ren. Toward robust material recognition for everyday objects. In *Proceedings of the British Machine Vision Conference*, pages 48.1–48.11. BMVA Press, 2011. http://dx.doi.org/10.5244/C.25.48.

[5] J. Duchi, E. Hazan, and Y. Singer. Adaptive subgradient methods for online learning and stochastic optimization. *J. Mach. Learn. Res.*, 12:2121–2159, July 2011.

[6] R. W. Fleming, W. C., and G. K. Perceptual qualities and material classes. *Vision Research*, (0), 2013.

[7] V. Franc, A. Zien, and B. Schölkopf. Support vector machines as probabilistic models. In *Proceedings of the 28th International Conference on Machine Learning, ICML 2011, Bellevue, Washington, USA, June 28 - July 2, 2011*, pages 665–672, 2011.

[8] P. V. Gehler, C. Rother, M. Kiefel, L. Zhang, and B. Sch"olkopf. Recovering intrinsic images with a global sparsity prior on reflectance. In J. Shawe-Taylor, R. S. Zemel, P. L. Bartlett, F. C. N. Pereira, and K. Q. Weinberger, editors, *NIPS*, pages 765–773, 2011.

[9] K. Jarrett, K. Kavukcuoglu, M. Ranzato, and Y. LeCun. What is the best multi-stage architecture for object recognition? In *Proc. International Conference on Computer Vision (ICCV'09)*. IEEE, 2009.

[10] Y. Jia, E. Shelhamer, J. Donahue, S. Karayev, J. Long, R. Girshick, S. Guadarrama, and T. Darrell. Caffe: Convolutional architecture for fast feature embedding. *arXiv preprint arXiv:1408.5093*, 2014.

[11] S. Karayev, A. Hertzmann, H. Winnemoeller, A. Agarwala, and T. Darrell. Recognizing image style. *CoRR*, abs/1311.3715, 2013.

[12] A. Krizhevsky, I. Sutskever, and G. E. Hinton. Imagenet classification with deep convolutional neural networks. In F. Pereira, C. Burges, L. Bottou, and K. Weinberger, editors, *Advances in Neural Information Processing Systems 25*, pages 1097–1105. Curran Associates, Inc., 2012.

[13] Y. Le Cun, L. Jackel, B. Boser, J. Denker, H. Graf, I. Guyon, D. Henderson, R. Howard, and W. Hubbard. Handwritten digit recognition: applications of neural network chips and automatic learning. *Communications Magazine, IEEE*, 27(11):41–46, Nov 1989.

[14] C. Liu, L. Sharan, E. H. Adelson, and R. Rosenholtz. Exploring features in a Bayesian framework for material recognition. In *CVPR*, pages 239–246. IEEE, 2010.

[15] L. Liu, P. Fieguth, G. Kuang, and H. Zha. Sorted random projections for robust texture classification. In *Computer Vision (ICCV), 2011 IEEE International Conference on*, pages 391–398, Nov 2011.

[16] S. J. Pan and Q. Yang. A survey on transfer learning. *IEEE Trans. on Knowl. and Data Eng.*, 22(10):1345–1359, Oct. 2010.

[17] A. S. Razavian, H. Azizpour, J. Sullivan, and S. Carlsson. CNN features off-the-shelf: an astounding baseline for recognition. *CoRR*, abs/1403.6382, 2014.

[18] J. Schmidhuber. Multi-column deep neural networks for image classification. In *Proceedings of the 2012 IEEE Conference on Computer Vision and Pattern Recognition (CVPR)*, CVPR '12, pages 3642–3649, Washington, DC, USA, 2012. IEEE Computer Society.

[19] G. Schwartz and K. Nishino. Visual material traits: Recognizing per-pixel material context. In *Computer Vision Workshops (ICCVW), 2013 IEEE International Conference on*, pages 883–890, Dec 2013.

[20] L. Sharan. Material perception: What can you see in a brief glance? *Journal of Vision*, 2009.

[21] L. Sharan, C. Liu, R. Rosenholtz, and E. H. Adelson. Recognizing materials using perceptually inspired features. volume 103, 2013.

[22] N. Srivastava, G. Hinton, A. Krizhevsky, I. Sutskever, and R. Salakhutdinov. Dropout: A simple way to prevent neural networks from overfitting. *Journal of Machine Learning Research*, 15:1929–1958, 2014.

[23] M. Weinmann, J. Gall, and R. Klein. Material classification based on training data synthesized using a btf database. In *Computer Vision - ECCV 2014 - 13th European Conference, Zurich, Switzerland, September 6-12, 2014, Proceedings, Part III*, pages 156–171. Springer International Publishing, 2014.

[24] G. Ye, E. Garces, Y. liu, Q. Dai, and D. Gutierrez. Intrinsic Video and Applications. *ACM Trans. Graph. (SIGGRAPH)*, 33(4), 2014.